\documentclass[12pt,a4paper]{article}

\usepackage{lineno, hyperref, multirow, amsmath, amssymb, color, mathtools, ulem}
\usepackage[a4paper]{geometry}
\usepackage[ruled]{algorithm2e}

\providecommand{\keywords}[1]
{
  \small	
  \textbf{\textit{Keywords---}} #1
}

\begin{document}

\title{Harmonic Convolutional Networks based on Discrete Cosine Transform}

\author{Matej Ulicny$^1$, Vladimir A. Krylov$^2$, Rozenn Dahyot$^3$}
\date{
    ADAPT Centre\\
    $^1$School of Computer Science \& Statistics, Trinity College Dublin, Ireland\\
    $^2$School of Mathematical Sciences, Dublin City University, Ireland\\
    $^3$Department  of Computer Science, Maynooth University, Ireland\\
}

\maketitle

\begin{abstract}
\noindent Convolutional neural networks (CNNs) learn filters in order to capture local correlation patterns in feature space. 
We propose to learn these filters as combinations of preset spectral filters defined by the Discrete Cosine Transform (DCT).
Our proposed DCT-based harmonic blocks  replace conventional convolutional layers to produce partially or fully harmonic versions of new or existing CNN architectures. 
Using DCT energy compaction properties, we demonstrate how the harmonic networks can be efficiently compressed  by truncating high-frequency information in harmonic blocks thanks to the redundancies in the spectral domain.
We report extensive experimental validation demonstrating benefits of the introduction of harmonic blocks into state-of-the-art CNN models in image classification, object detection and semantic segmentation applications.
\end{abstract}

\keywords{
Harmonic Network, Convolutional Neural Network, Discrete Cosine Transform, Image Classification, Object Detection, Semantic Segmentation
}

\section{Introduction}\label{sec:introduction}

CNNs have been designed to take advantage of implicit characteristics of natural images, specifically correlation in local neighborhood and feature equivariance.
Standard CNNs rely on learned convolutional filters hence finetuned to the data available. However, it can be advantageous to revert to preset filter banks: for instance, with limited training data ~\cite{Ulicny19}, using a  collection of preset filters can help in avoiding overfitting and in reducing the computational complexity of the system.
Scattering networks are an example of such networks with preset (wavelet based) filters  which have achieved state-of-the-art results in handwritten digit recognition and texture classification~\cite{Bruna13}.

\begin{figure}[!h]
\begin{center}
\begin{center}
   \includegraphics[width=0.27\linewidth]{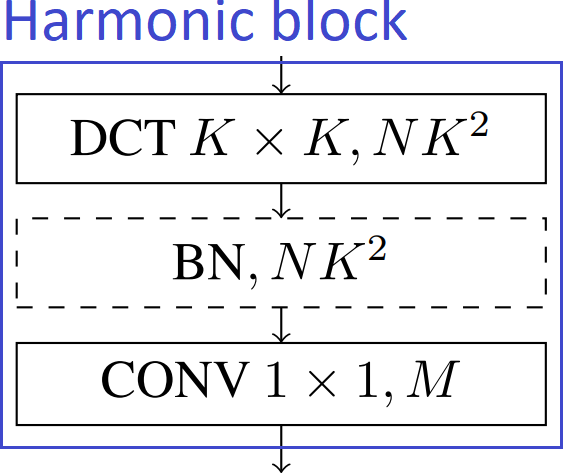}
   \includegraphics[width=0.72\linewidth]{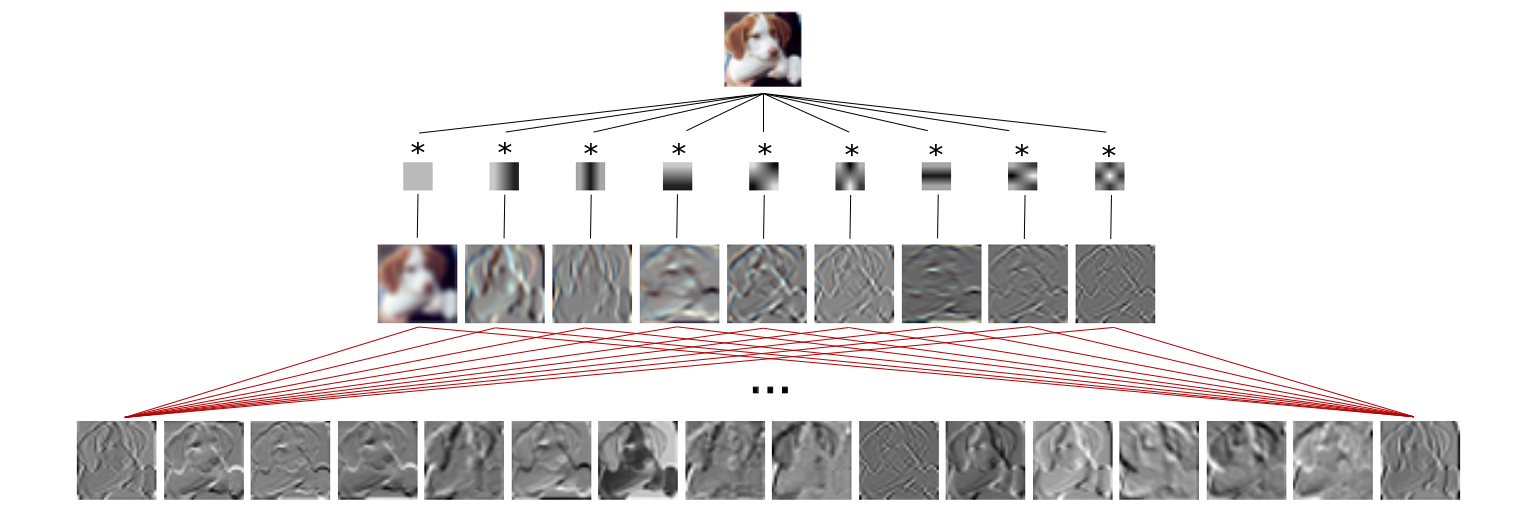}
\end{center}
\end{center}
\vspace{-.5\baselineskip}
\caption{Left: Design of the harmonic block. 
Boxes show operation type, size of filter (if applicable) and the number of output channels given the block filter size $K$, number of input channels $N$ and output channels $M$. Batch normalization (BN) block is optional. Right: Visualization of the harmonic block applied to an input layer.}
\label{fig:harmlayer}
\end{figure}

We propose instead to replace the standard convolutional operations in CNNs by harmonic blocks that learn the weighted sums of responses to the Discrete Cosine Transform (DCT) filters, see Fig.~\ref{fig:harmlayer}.
DCT has been successfully used for JPEG encoding to transform image blocks into spectral representations to capture the most information with a small number of coefficients.
Motivated by frequency separation and energy compaction properties of DCT, the proposed harmonic networks rely on combining responses of window-based DCT with a small receptive field.
Our method learns how to optimally combine spectral coefficients at every layer to produce a fixed size representation defined as a weighted sum of responses to DCT filters. The use of DCT filters allows one to represent and regularize filter parameters directly in the DCT domain and easily address the task of model compression.
Other works that propose convolutional filters decomposition to particular basis functions~\cite{Qiu18,Tayyab19} have predominantly focused on network compression. In our study we demonstrate that prior information coming from well chosen filter basis can not only be used to compress but also speeds up training convergence and improves performance.

Based on our earlier works~\cite{Ulicny19b,Ulicny19},  
this paper  contributions are as follows. 
First we demonstrate that the theoretical computational overheads of the optimised formulation of a harmonic block are minimal (experimentally, within 3-7\%) whereas the memory footprint requirements are comparable to those of the benchmark architecture based on standard convolutional blocks (and are lower if harmonic blocks undergo compression). Second, we substantially expand experimental validation to demonstrate a consistent increase in performance due to the use of harmonic blocks. Specifically, on the small NORB dataset we achieve state-of-the-art results and demonstrate how DCT-based harmonic blocks allow one to efficiently generalise to unseen lighting conditions. We further report quantitative as well as qualitative results of application of harmonic blocks to a representative variety of vision tasks: object detection and instance/semantic segmentation. We observe a consistent average improvement of 1\% AP on these tasks, which demonstrates the practical appeal of using harmonic networks.
Section ~\ref{sec:soa} presents the relevant background research to   our harmonic network formulation (Sec.~\ref{sec:method}).
It is extensively validated against state-of-the-art alternatives  for image classification (Sec.~\ref{sec:experiments1}), for object detection, instance and semantic segmentation (Sec.~\ref{sec:experiments.detection}).
All our architectures in the reported results are denoted as \textit{Harm};
the PyTorch implementations for our harmonic networks are publicly available at \url{https://github.com/matej-ulicny/harmonic-networks}.

\section{Related work}
\label{sec:soa}

\subsection{DCT \& CNNs }

Networks trained on DCT coefficients are frequently used in forensics for detection of tampered parts in images. These parts are assumed to have different distribution of DCT coefficients from the rest of the image.
A common practice is to classify histograms of preselected JPEG-extracted DCT coefficients by 1-D or 2-D convolutional network~\cite{Zheng19}.
A number of studies have also investigated the use of spectral image representations for object and scene recognition. 
DCT features from an entire image were used to train Radial Basis Function Network for face recognition~\cite{Er05}. 
A DCT-based scene descriptor was used together with a CNN classifier~\cite{Farinella15}.
A significant convergence speedup and case-specific accuracy improvement have been achieved by applying DCT transform to early stage learned feature maps in shallow CNNs~\cite{Ghosh16} whereas the later stage convolutional filters were operating on a sparse spectral feature  representation. 
In~\cite{Ulicny17,Gueguen18} it was demonstrated how DCT coefficients can be efficiently used to train CNNs for classification, where the DCT coefficients can be computed or taken directly from JPEG image format. 
\textit{Wang et al.}~\cite{Wang16b} compresses CNN filters by separating cluster centers and residuals in their DCT representation. Weights in this form were quantized and transformed via Huffman coding (used for JPEG compression) for limiting storage. Lastly, DCT image representation has been used for calculation of more informative loss function in generative learning~\cite{Atapour19}. 

\subsection{ Wavelets \& CNNs}

\paragraph{Wavelet networks} 
As an alternative to DCT, scattering networks~\cite{Bruna13} are built on complex-valued wavelets.
The scattering network has its filters designed to extract translation and rotation invariant representations.
It was shown to effectively reduce the input representation while preserving discriminative information for training CNN on image classification and object detection task~\cite{Oyallon18b} achieving performance comparable to deeper models. \textit{Williams et al.}~\cite{Williams16} have advocated image preprocessing with wavelet transform, but used different CNNs for each frequency subband. Wavelet filters were also used as a preprocessing method prior to NN-based classifier~\cite{Said16}.

\paragraph{Spectral based CNNs}
Other works have used wavelets in CNN computational graphs. 
Low-frequency components of the Dual-Tree Complex Wavelet transform were used in a noise suppressing pooling operator~\cite{Duan17}. 
\textit{Ripperl et al.} have designed a spectral pooling~\cite{Rippel15} based on Fast Fourier Transform and truncation of high-frequency coefficients. 
The pooled features were recovered with Inverse Discrete Fourier Transform, thus the CNN still operates in the spatial domain.
They also proposed to parameterise filters in the Fourier domain to decrease their redundancy and speed up the convergence when training the network.

A Wavelet Convolutional Network proposed by \textit{Lu et al.}~\cite{Lu18} learns from both spatial and spectral information that is decomposed from the first layer features. The higher-order coefficients are concatenated along with the feature maps of the same dimensionality. 
However, contrary to our harmonic networks, Wavelet CNNs decompose only the input features and not the features learned at intermediate stages. 
Robustness to object rotations was addressed by modulating learned filters by oriented Gabor filters~\cite{Luan18}. \textit{Worrall et al.} incorporated complex circular harmonics into CNNs to learn rotation equivariant representations~\cite{Worrall17}. Similarly to our harmonic block, the structured receptive field block~\cite{Jacobsen16} learns new filters by combining fixed filters, a set of Gaussian derivatives with considerably large spatial extent. Additionally, an orthogonal set of Gaussian derivative bases of small spatial extend have been used by \textit{Kobayashi} to express convolutional filters~\cite{Kobayashi18}. DCFNet~\cite{Qiu18} expresses filters by truncated expansion of Fourier-Bessel basis, maintaining accuracy of the original model while reducing the number of parameters. 


\section{Harmonic Networks}\label{sec:method}

A convolutional layer extracts correlation of input patterns with locally applied learned filters. The idea of convolutions applied to images stems from the observation that pixels in local neighborhoods of natural images tend to be strongly correlated. In many image analysis applications, transformation methods are used to decorrelate signals forming an image~\cite{Wang12}. In contrast with spatial convolution with learned kernels, this study proposes feature learning by weighted combinations of responses to predefined filters. The latter extracts harmonics from lower-level features in a region. The use of well selected predefined filters allows one to reduce the impact of overfitting and decrease computational complexity. We focus here on the use of DCT as the underlying transformation.

\subsection{Discrete Cosine Transform} \label{sec:dct}

DCT is an orthogonal transformation method that decomposes an image to its spatial frequency spectrum. A 2D signal is expressed as a sum of sinusoids with different frequencies. The contribution of each sinusoid towards the whole signal is determined by its coefficient calculated during the transformation.
DCT is also a separable transform and due to its energy compaction properties on natural images~\cite{Wang12} it is commonly used for image and video compression in widely used JPEG and MPEG formats. Note that Karhunen-Lo{\`e}ve transform (KLT) is considered to be optimal in signal decorrelation, however it transforms signal via unique basis functions that are not separable and need to be estimated from the data. On locally correlated signals such as natural images DCT was shown to closely approximate KLT~\cite{Wang12}.

We use the most common DCT formulation, noted DCT-II,  computed on a 2-dimensional grid of an image $X$ of size $A \times B$ representing the image patch with 1 pixel discretisation step:
\begin{equation} \label{eq:dct}
Y_{u,v} = \sum_{x=0}^{A-1} \sum_{y=0}^{B-1}  \sqrt{\frac{\alpha_u}{A}} \sqrt{\frac{\alpha_v}{B}} X_{x,y} \cos{\left[\frac{\pi}{A} \left(x+\frac{1}{2}\right)u\right]} \cos{\left[\frac{\pi}{B}\left(y+\frac{1}{2} \right)v\right]}.
\end{equation}
$Y_{u,v}$ is  the DCT coefficient  of the input $X$ using a sinusoid with horizontal and vertical frequencies noted $u$  and $v$ respectively. Basis functions are typically normalized with factors $\alpha_{u}=1$ (resp. $\alpha_{v}=1$) when $u=0$ (resp. when $v=0$)   and $\alpha_{u}=2 $ (resp. $\alpha_{v}=2 $) otherwise to ensure their orthonormality. 

It is worth noting~\cite{Ulicny19}, that the sine transform of the signal $X$ with $N$ values at frequency $k$ is equivalent to the cosine transform of the image shifted by $N\left(1+4z\right)/2k$ pixels, $z \in \mathbb{Z}$:
\begin{equation} \label{eq:dct_shifted_data}
\sum_{n=0}^{N-1} X_n \sin{\left[\frac{\pi}{N} \left(n+\frac{1}{2}\right)k\right]}
=\sum_{n=0}^{N-1}{X_{n+\frac{N(1+4z)}{2k}} \cos{\left[ \frac{\pi}{N} \left( n+\frac{1}{2} \right) k \right]}}.
\end{equation}
Hence by applying DCT with a certain stride (effectively resulting in {\it overlapping} DCT transform) it is possible to obtain a feature representation as rich as that obtained employing the full Fourier transform~\cite{Ulicny19}.

\subsection{Harmonic blocks}

We propose the harmonic block  to replace a conventional convolutional operation hence relying on processing the data in two stages (see Fig.~\ref{fig:harmlayer}). Firstly, the input features undergo harmonic decomposition using window-based DCT.
In the second stage, the transformed signals are combined by learned weights. 
The fundamental difference from standard convolutional network is that the optimization algorithm is not searching for filters that extract spatial correlation, rather learns the relative importance of preset feature extractors (DCT filters) at multiple layers.

Harmonic blocks are integrated as a structural element in the existing or new CNN architectures. We thus design harmonic networks that consist of one or more harmonic blocks and, optionally, standard learned convolutions and fully-connected layers, as well as any other structural elements of a neural net.
Spectral decomposition of input features into block-DCT representation is implemented as a convolution with DCT basis functions. A 2D kernel with size $K \times K$ is constructed for each basis function, comprising a filter bank of depth $K^2$, which is separately applied to each of the input features. Convolution with the filter bank isolates coefficients of DCT basis functions to their exclusive feature maps, creating a new feature map per each channel and each frequency considered. The number of operations required to calculate this representation can be minimized by decomposing 2D DCT filter into two rank-1 filters and applying them as separable convolution to rows and columns sequentially. 

Each feature map $h^l$ at depth $l$ is computed as a weighted linear combination of DCT coefficients across all input channels $N$:
\begin{equation} \label{eq:feature}
  h^l = \sum_{n=0}^{N-1}{\sum_{u=0}^{K-1}{\sum_{v=0}^{K-1}{w^{l}_{n,u,v}\psi_{u,v}* *\, h^{l-1}_n}}}
\end{equation}
where $\psi_{u,v}$ is a $u,v$ frequency selective DCT filter of size $K \times K$, $**$ the 2-dimensional convolution operator and $w^{l}_{n,u,v}$ is learned weight for $u,v$ frequency of the $n$-th feature. The linear combination of spectral coefficients is implemented via a convolution with $1\times 1$ filter that scales and sums the features, see Fig.~\ref{fig:harmlayer}.
In our implementation we use a fixed collection of DCT bases. Specifically, if we are to replace a $K\times K$ convolution layer, the DCT filter bank $\left\{\psi_{u,v} \in \mathbb{R}^{K\times K}; u,v \in \mathbb{N}; 0 \leq u,v < K\right\}$ has filters defined for every filter coordinate $x,y$ as given in Eq.~\ref{eq:dct}.
Since the DCT is a linear transformation, backward pass through the transform layer is performed similarly to a backward pass through a convolution layer. Harmonic blocks are designed to learn the {\it same} number of parameters as their convolutional counterparts. Such blocks can be considered a special case of depth-separable convolution with predefined spatial filters. 

DCT is distinguished by its energy compaction capabilities which typically results in higher filter responses in lower frequencies. The behaviour of relative loss of high frequency information can be efficiently handled by normalizing spectrum of the input channels. This can be achieved via batch normalization that adjusts per frequency mean and variance prior to the weighted combination. The spectrum normalization transforms Eq.~\eqref{eq:feature} into: 
\begin{equation} \label{eq:normalized_feature}
  h^l = \sum_{n=0}^{N-1}{\sum_{u=0}^{K-1}{\sum_{v=0}^{K-1}{w^l_{n,u,v}\frac{\psi_{u,v}** h^{l-1}_n - \mu^l_{n,u,v}}{\sigma^l_{n,u,v}}}}},
\end{equation}
with parameters $\mu^l_{n,u,v}$ and $\sigma^l_{n,u,v}$ estimated per input batch.

\subsection{Harmonic Network Compression} \label{sec:method.subsample}

\begin{figure}[t]
\begin{center}
   \includegraphics[width=.8\linewidth]{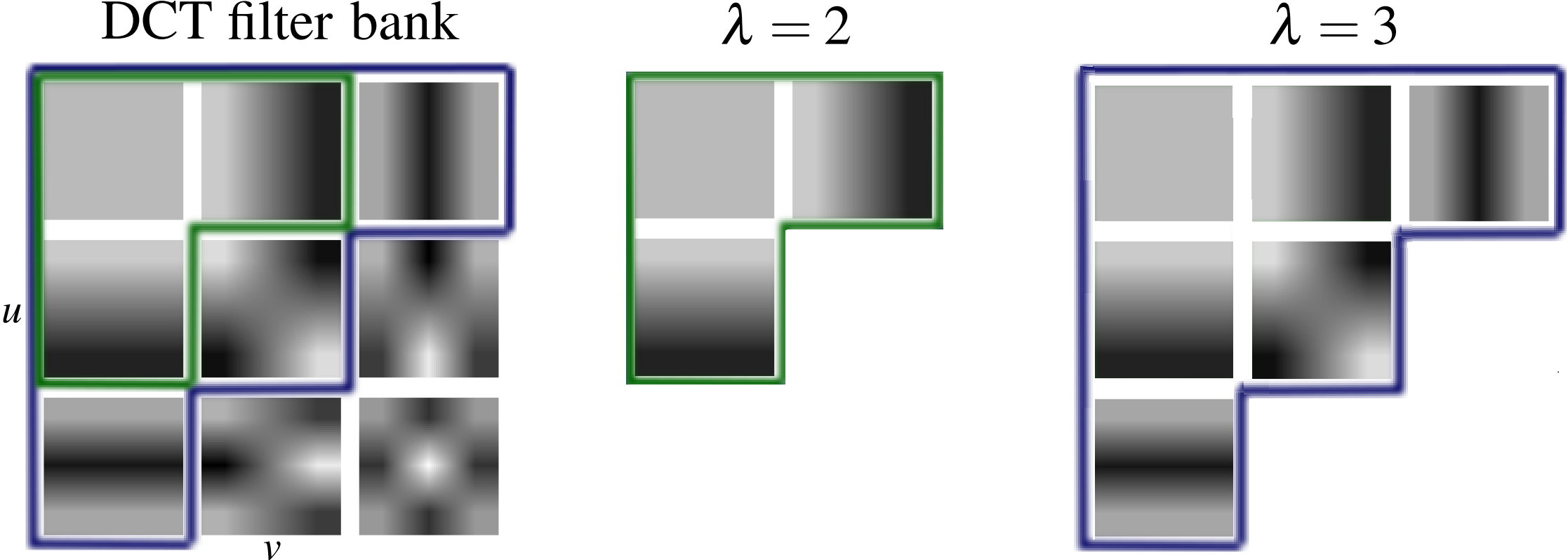}
\end{center}
\vspace{-.5\baselineskip}
   \caption{ $3\times 3$ DCT filter bank employed in the harmonic networks and its compression.}
\label{fig:filters}
\end{figure}

The JPEG compression encoding relies on stronger quantisation of higher frequency DCT coefficients. This is motivated by the human visual system which often prioritises low frequency information over high frequencies. We propose to employ similar idea in the harmonic network architecture. Specifically, we limit the visual spectrum of harmonic blocks to only several most informative low frequencies, which results in a reduction of number of parameters and operations required at each block. The coefficients are (partially) ordered by their relative importance for the visual system in triangular patterns starting at the most important zero frequency at the top-left corner, see Fig.~\ref{fig:filters}. We limit the spectrum of considered frequencies by hyperparameter $\lambda$ representing the number of levels of coefficients included perpendicularly to the main diagonal direction starting from zero frequency: DC only for $\lambda=1$, three coefficients used for $\lambda=2$, and six coefficients used for $\lambda=3$.
Fig.~\ref{fig:filters} illustrates filters used at various levels assuming a $3\times 3$ receptive field.
Thus, reformulating convolutional layers as harmonic allows one  to take advantage of this natural approach to model compression, and in doing also introduce additional regularization into the model.

When compressing harmonic networks with multiple harmonic blocks a question what is the appropriate subset of coefficients arises. We propose 3 principles of selecting these subsets: uniform, progressive and adaptive. 

\noindent\textbf{Uniform selection} is the most simple compression approach that uses the same $\lambda$ at every layer. A pitfall of this method is that a specific subset of basis functions might not explain well enough filter banks at some layers, but might have some bases redundant when representing filters in other layers.

\noindent\textbf{Progressive selection} omits higher number of frequencies in deeper layers. We investigate two strategies for progressive coefficient selection. The first strategy employs the same subset of coefficients in all harmonic blocks applied to feature maps of a particular size, but this subset shrinks with the size of feature maps (smaller maps in deeper layers). 
The second strategy relies on the selection of compression level based on the depth of the layer. Specifically, the compression level is estimated as $\lambda_{progr} = \max(\alpha,\min(2K-1,\lfloor{T/Depth}\rfloor))$, where $\alpha$ is the lowest allowed value of $\lambda$, considering $\alpha\in\left\{1, 2\right\}$, $K$ the filter size ($\lambda = 2K-1$ corresponds to no compression), and $T$ a constant that sets a linear relation between $\lambda$ and the depth of a particular layer and controls the overall compression rate when $T\in\left<\alpha,\left(2K-1\right)\times Depth\right>$.
    
\noindent\textbf{Adaptive selection}: compression is estimated adaptively for each layer; a basis is excluded if proportion of the $L1$ norm of its corresponding weights compared to the norms of the other bases in the same layer is lower than a threshold $T$. Specifically, if $\left| w_{i,j} \right|_1 / \sum_{u,v=0}^{K-1}{\left| w_{u,v} \right|_1}<T$ then the coefficient is truncated. It should be noted that this approach, unlike the previous ones, needs the full model to be optimized prior to compression.

The empirical impact of harmonic model compression is  investigated experimentally in more detail in Sections \ref{sec:experiments1} and \ref{sec:experiments.detection}.

\subsection{Computational Requirements} \label{sec:methodrequirements}

Harmonic blocks are designed to learn the same number of parameters as their convolutional counterparts. Requirements for the DCT transform scale linearly with the number of input channels and result in a modest increase to the theoretical number of operations. Standard convolutional layer used in many popular architectures that has $N$ input and $M$ output channels with a kernel size $K\times K$ learns $NMK^2$ parameters and performs $NMK^2AB$ operations if the filter is applied $A$ and $B$ times in particular directions. Harmonic block with $K^2$ transformation filters of size $K\times K$ upsamples representation to $NK^2$ features and then learns one weight for each upsampled-output feature pair hence $NK^2M$ weights. Transformation of an $A\times B$ feature set costs $NK^2K^2AB$ on top of weighted combination $NK^2MAB$ that matches the number of multiply-add operations of $K \times K$ convolution. The total number of operations is thus $NK^2AB\left(M+K^2\right)$. The theoretical number of multiply-add operations over the standard convolutional layer increases by a factor of ${K^2}/{M}$. If we assume truncated spectrum (use of $\lambda \leq K$) given by $P=\lambda(\lambda+1)/2$ filters, proportion of operations becomes $P/K^2+P/M$.

While keeping the number of parameters intact, a harmonic block requires additional memory during training and inference to store transformed feature representation. In our experiments with WRN models (Sec.\ref{sec:experiments.cifar}), the harmonic network trained with full DCT spectrum requires almost 3 times more memory than the baseline.
This memory requirement can be reduced by using the DCT spectrum compression.

Despite the comparable theoretical computational requirements, the run time of harmonic networks is larger compared to the baseline models, at least twice slower (on GPU) in certain configurations. This effect is due to generally less efficient implementation of separable convolution and the design of harmonic block that replaces a single convolutional layer by a block of 2 sequential convolutions (with individual harmonic filters and 1x1 convolution). Most blocks do not need BN between the convolutions and thus represent a combined linear transformation. The associativity property of convolutions allows one to reformulate the standard harmonic block defined above so that the DCT transform and linear combination can be effectively merged into a single linear operation:
\begin{equation}
\label{eq:filter_decomp}
h^l = \sum_{n=0}^{N-1}{ \sum_{u=0}^{K-1}{ \sum_{v=0}^{K-1}{w_{n,u,v} \left( \psi_{u,v} ** h_n^{l-1} \right) }}} 
= \sum_{n=0}^{N-1}{ \left( \sum_{u=0}^{K-1}{ \sum_{v=0}^{K-1}{ w_{n,u,v} \psi_{u,v} }} \right) ** h_n^{l-1} } 
\end{equation}
In other words, equivalent features can be obtained by factorizing filters as linear combinations of DCT basis functions. We thus propose a faster Algorithm~\ref{alg:mem_eff_harm_block} that is a more memory efficient alternative to the standard two-stage harmonic block formulation and uses dense convolution. 

\begin{algorithm}[t]
 \KwIn{$h^{l-1}$}
 {Define updates} $g \in \mathcal{R}^{M \times N \times K \times K}$\;
 \For{$m \in \left\{ 0..M-1 \right\}$}{
  \For{$n \in \left\{ 0..N-1 \right\}$}{
   $g^l_{m,n} \leftarrow \sum_{u=0}^{K-1}{\sum_{v=0}^{K-1}{w_{m,n,u,v}\ \psi_{u,v}}}$\;
  }
 }
 $h^l \leftarrow g^l**h^{l-1}$\;
 \KwOut{$h^l$}
 \caption{Memory efficient harmonic block} 
 \label{alg:mem_eff_harm_block}
\end{algorithm}
The Algorithm~\ref{alg:mem_eff_harm_block} overhead in terms of multiply-add operations with respect to the standard convolutional layer is only $K^2/AB$, where the input image size for the block is $A\times B$. The experimental performance of the algorithm is evaluated in Section~\ref{sec:experiments.cifar}.

\section{Image Classification } 
\label{sec:experiments1}

The performance of the harmonic networks is assessed for image classification  on small (NORB, Sec. \ref{sec:experiments.norb}), medium (CIFAR-10 and CIFAR-100, Sec \ref{sec:experiments.cifar}) and large (ImageNet-1K, Sec. \ref{sec:experiments.imagenet}) scale datasets.

\subsection{Small NORB dataset} \label{sec:experiments.norb}

The small NORB dataset~\cite{Lecun04} is a synthetic set of $96\times 96$ binocular images of 50 toys sorted into 5 classes (four-legged animals, human figures, airplanes, trucks, and cars), captured under different lighting and pose conditions (i.e. 18 angles, 9 elevations and 6 lighting conditions induced by combining different light sources). Training  and test sets used in our experiments are retained original~\cite{Lecun04}.
We show first that harmonic networks outperform standard and state-of-the-art CNNs in both accuracy and compactness (c.f. Section \ref{sec:norb:comparison}) and also illustrate how Harmonic networks can be naturally resilient to unseen illumination changes without resorting to using data augmentation (Sec. \ref{sec:norb:illumination}).

\subsubsection{Comparisons CNN vs. Harmonic Nets}
\label{sec:norb:comparison}

\vskip .1cm
\noindent{\bf Baseline architectures.} 
Our baseline CNN2 consists of 2 convolution and 2 fully-connected layers. Features are subsampled by convolution with stride and overlapping max-pooling. All hidden layer responses are batch normalized and rectified by ReLU. We also use a slightly deeper network CNN3 with an additional convolutional layer preceding the first pooling. Details of the architectures are summarised in Table~\ref{tab:norb_nn}. 

\noindent{\bf Optimisation.}
The baseline CNNs are trained with stochastic gradient descent for 200 epochs with momentum 0.9 and weight decay 0.0005. The initial learning rate 0.01 is decreased by factor 10 every 50 epochs. The network is trained with batches of 64 stereo image pairs, each pair is zero-padded 5 pixels and a random crop of 96$\times$96 pixels is fed to the network.

\begin{table}[!t]
\tabcolsep = 1.0mm
\begin{center}
\caption{Models used in NORB experiments.
Convolution and harmonic operation are denoted as \{conv, harm\}~M,K$\times$K/S with M output features, kernel size K and stride S; similarly for pooling~K$\times$K/S and fully connected layers~fc~M.}
\label{tab:norb_nn}
\vspace{0.3\baselineskip}
\footnotesize
\begin{tabular}{cccccc}
\hline
\textbf{Resol.} & \textbf{CNN2} & \textbf{CNN3} & \textbf{Harm-CNN2}& \textbf{Harm-CNN3 }& \textbf{Harm-CNN4}\\
\hline
96x96 & conv 32, 5x5/2 & conv 32, 5x5/2 & harm 32, 4x4/4 & harm 32, 4x4/4 & harm 32, 4x4/4\\
48x48 & pool 3x3/2 & conv 64, 3x3/2 & - & - & - \\
24x24 & conv 64, 3x3/2 & pool 2x2/2 & harm 64, 3x3/2 & harm 64, 3x3/2& harm 64, 3x3/2 \\
12x12 & pool 3x3/2 & conv 128, 3x3/2 &  pool 3x3/2&pool 3x3/2&pool 3x3/2\\
6x6 & fc 1024 & pool 2x2/2 & fc 1024 & harm 128, 3x3/2 & harm 128, 3x3/2 \\
3x3 & - & fc 1024 & - & fc 1024 & harm 1024, 3x3/3 \\
1x1 & dropout 0.5 & dropout 0.5 & dropout 0.5& dropout 0.5& dropout 0.5\\
1x1 & fc 5 & fc 5 & fc 5& fc 5& fc 5\\
\hline
\end{tabular}
\end{center}
\end{table}

\noindent{\bf Harmonic Networks architectures.} Several versions of harmonic networks are considered (Tab. \ref{tab:norb_nn}), by substituting the first, first two or all three of CNN2 and CNN3 convolution layers by harmonic blocks. Furthermore, the first fully-connected layer can be transformed to a harmonic block taking global DCT transform of the activations. The first harmonic block uses 4$\times4$ DCT filters, the further blocks mimic their convolutional counterparts.

\begin{figure}[t]
\begin{center}
   \includegraphics[width=.6\linewidth]{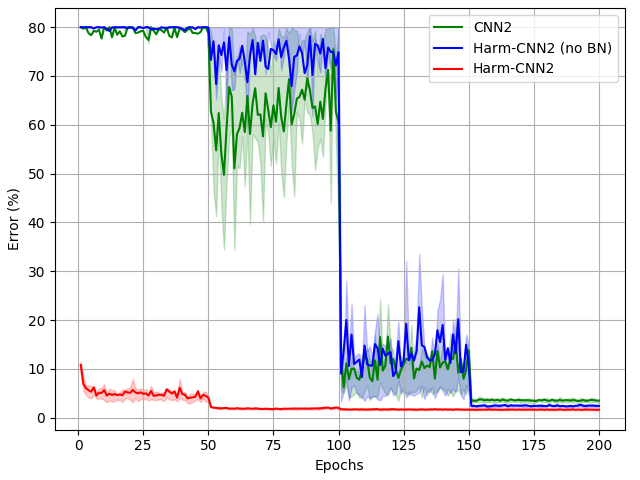}
\end{center}
\vspace{-1.5\baselineskip}
\caption{Mean classification error on small NORB test set. Weak generalization of CNN (green) and harmonic network (blue) is observed during the early stages of training. Filled areas (best seen in color) show 50\% empirical confidence intervals from 20 runs. Batch normalization of DCT spectrum (first block) significantly speeds up convergence of harmonic network (red).}
\label{fig:norb_convergence}
\end{figure}

\vskip .1cm
\noindent{\bf Performance evaluation.} The baseline CNN architecture shows poor generalization performance in early stages of training, see Fig.~\ref{fig:norb_convergence}. Baseline CNN2 achieved mean error 3.48\%$\pm$0.50 from 20 trials, while CNN2 utilizing harmonic blocks without explicit normalization of harmonic responses exhibits similar behavior resulting in lower mean error of 2.40\%$\pm$0.39. Normalizing DCT responses at the first block prevents harmonic network from focusing too much on pixel intensity, allows using 10$\times$ higher learning rate, significantly speeds up convergence, improves performance and stability. 
All variants of the harmonic network perform comparably.
Particularly we observe the overlapping average pooling to work well in combination with harmonic blocks. The best result was obtained by the Harm-CNN4 model with 4 harmonic blocks (the latter replaces the fully-connected layer), misclassifying only 1.10\%$\pm$0.16 of test samples.

\vskip .1cm
\noindent{\bf Comparison with state-of-the-art.} Table~\ref{tab:norb_state_of_art} shows that these results surpass the best previously reported error rate for this dataset to the best of our knowledge. The capsule network~\cite{Hinton18} claims 1.4\% error rate, however estimated under a different evaluation protocol.

\begin{table}[h]
\begin{center}
\caption{Comparison with the state-of-the-art on small NORB dataset, showing the proposed method outperforms other reported results.} 
\label{tab:norb_state_of_art}
\vspace{0.3\baselineskip}
\tabcolsep = 1mm
\begin{tabular}{ l c c c c }
\hline
\textbf{Method} & \textbf{Parameters} $\downarrow$ & \textbf{Error \%} $\downarrow$ \\
\hline
CNN3 & 1.28M & 3.43 $\pm$ 0.31 \\
CapsNet~\cite{Hinton18} multi-crop & 310K & 1.4* \\
\hline
Harm-CNN2 & 2.39M & 1.56 $\pm$ 0.18 \\
Harm-CNN3 & 1.28M & 1.15 $\pm$ 0.22 \\
Harm-CNN4 & 1.28M & \textbf{1.10 $\pm$ 0.16} \\
\hline
\end{tabular}\\[.05cm]
*score reported by the authors of the corresponding paper.
\end{center}
\end{table}

\vskip .1cm
\noindent{\bf Harmonic network compression.}
We further proceed to designing a very compact version of Harm-CNN4 (cf. Table~\ref{tab:norb_state_of_art}). To do so, the fully connected layer is reduced to only 32 neurons and the dropout is omitted. The modified network reaches 1.17\%$\pm$0.20 error and has 131k parameters. When applying compression with $\lambda=K$ (i.e. 3 for $3\times3$ filters) the network reaches 1.34\%$\pm$0.21 and requires less than 88k parameters.
By applying $\lambda=K-1$ the total count is less than 45k parameters and error of 1.64\%$\pm$0.22 is achieved on the test set, in contrast with small capsule network~\cite{Hinton18} with 68k parameters scoring a higher error of 2.2\%.

\subsubsection{Harmonic Networks for illumination changes}
\label{sec:norb:illumination}

Spectral representation of input data has a few interesting properties. Average feature intensity is captured into DC coefficient, while other (AC) coefficients capture the signal structure. DCT representation has been successfully used~\cite{Er05} to build illumination invariant representation. This gives us strong motivation to test illumination invariance properties of harmonic networks and to compare them with standard CNNs. 
Objects in the small NORB dataset are normalized and are presented with their shadows over a uniform background. The six lighting conditions are obtained by various combination of up to 4 fixed light sources at different positions and distances from the objects.

Usual approaches to reduce sensitivity to lighting conditions include image standardization with ZCA whitening or illumination augmentation. 
Brightness and contrast manipulations encourage the network to focus on features that are independent of the illumination.
Contrary to these methods in our approach we achieve the same effect by removing the filter corresponding to the DC component from the first harmonic block. Such network is invariant towards global additive changes in pixel intensity by definition.  

\vskip .1cm
\noindent{\bf Set-up.} The dataset is split into 3 parts based on lighting conditions during image capturing: the bright images (conditions 3,5) dark images (cond. 2,4) and images under standard lighting conditions (cond. 0,1). The models are trained (w/wo data augmentation) only on data from one split and tested on images from the other two splits that contain unseen lighting conditions.

\begin{table}[!h]
\begin{center}
\caption{Means of classification error over 10 runs on NORB test images captured in unseen illumination conditions. Harmonic networks improve classification error of CNN by 5--16\%. \label{tab:illum_inv}}
\vspace{0.3\baselineskip}
\begin{tabular}{ l c c c c }
\hline
\textbf{Augmentation} & \multicolumn{2}{c}{\textbf{None}} & \multicolumn{2}{c}{\textbf{Brightness \& contrast}} \\
\textbf{Lighting Condition} & \textbf{CNN} $\downarrow$& \textbf{Harmonic}$\downarrow$ & \textbf{CNN} $\downarrow$& \textbf{Harmonic}$\downarrow$ \\
\hline
Bright & 26.3$\pm$2.6 & 10.2$\pm$0.4 & 17.6$\pm$0.7 & 9.4$\pm$0.7\\
Standard & 30.2$\pm$1.8 & 18.0$\pm$1.9 & 22.5$\pm$1.1 & 15.1$\pm$1.1 \\
Dark & 31.2$\pm$1.8 & 18.9$\pm$1.2 & 20.1$\pm$1.3 & 14.5$\pm$1.4 \\
\hline
\end{tabular}
\end{center}
\end{table}

\vskip .1cm
\noindent{\bf Performance evaluation.}
Classification errors of the best CNN and harmonic network architectures on the test images (unseen illumination conditions) are reported in Table~\ref{tab:illum_inv}.
Harmonic networks consistently achieve lower error under various unseen lighting conditions in comparison to baseline CNNs, with and without random brightness and contrast (only for dark images) augmentation.

\subsection{CIFAR-10/100 datasets} \label{sec:experiments.cifar}

The second set of experiments is performed on popular benchmark datasets of small natural images CIFAR-10 and CIFAR-100.

\vskip .1cm
\noindent{\bf Baseline.} For experiments on CIFAR datasets we adopt WRNs~\cite{Zagoruyko16} with 28 layers and width multiplier 10 (WRN-28-10) as the main baseline. 
Model design and training procedure are kept as in the original paper. Harmonic WRNs are constructed by replacing convolutional layers by harmonic blocks with the same receptive field, preserving batch normalization and ReLU activations in their original positions after every block.

\vskip .1cm
\noindent{\bf Results.} We first investigate whether the WRN results can be improved if only trained on spectral information, replacing only the first convolutional layer (denoted as Harm1-WRN). The network learns more useful features if the RGB spectrum is explicitly normalized by integrating the BN block as demonstrated in Fig.~\ref{fig:harmlayer}, surpassing the classification error of the baseline network on both CIFAR-10 and CIFAR-100 datasets, see Table~\ref{tab:cifar_spec}. We then construct a fully harmonic WRN (Harm-WRN) by replacing all convolutional layers with harmonic blocks.
This harmonic network also outperforms the baseline WRN, see Table~\ref{tab:cifar_spec}.

\begin{table}[!t]
\begin{center}
\caption{Settings and median error rates (\%) out of 5 runs achieved by WRNs and their harmonic modifications on CIFAR datasets. Number of parameters reported for CIFAR-10.} 
\label{tab:cifar_spec}
\vspace{0.3\baselineskip}
\tabcolsep = 1.7mm
\small
\begin{tabular}{ l c c c c c c c }
\hline
\textbf{Method} & \textbf{Dropout}  & \textbf{Param.} $\downarrow$ & \textbf{CIFAR-10} $\downarrow$ & \textbf{CIFAR-100} $\downarrow$\\
\hline
WRN-28-10~\cite{Zagoruyko16} & \checkmark& 36.5M & 3.91 & 18.75 \\
Gabor CNN 3-28~\cite{Luan18} &  &  17.6M & \hspace{.4ex} 3.88* & \hspace{.4ex} 20.13* \\
\hline
Harm1-WRN-28-10 (no BN) & & 36.5M & 4.10 & 19.17 \\
Harm1-WRN-28-10 & & 36.5M & 3.90 & 18.80 \\
Harm1-WRN-28-10 & \checkmark & 36.5M & \textbf{3.64} & \textbf{18.57} \\
Harm-WRN-28-10 & \checkmark & 36.5M & 3.86 & \textbf{18.57} \\
Harm-WRN-28-10, progr. $\lambda$  & & 15.7M & 3.93 & 19.04 \\
\hline
\end{tabular}\\[.05cm]
*scores reported by~\cite{Luan18}.
\end{center}
\end{table}

\begin{figure}[!b]
\begin{center}
   \includegraphics[width=.9\linewidth]{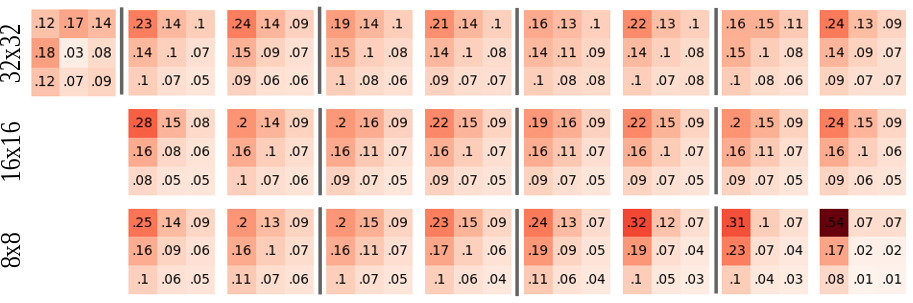}
\end{center}
\vspace{-.25\baselineskip}
\caption{Distribution of weights (averaged in each layer) assigned to DCT filters in the first harmonic block (left-most) and the remaining blocks in the Harm-WRN-28-10 model trained on CIFAR-10. Vertical lines separate the residual blocks.}
\label{fig:freq_heatmap}
\end{figure}

\begin{figure*}[!t]
\begin{center}
   \includegraphics[width=.49\linewidth]{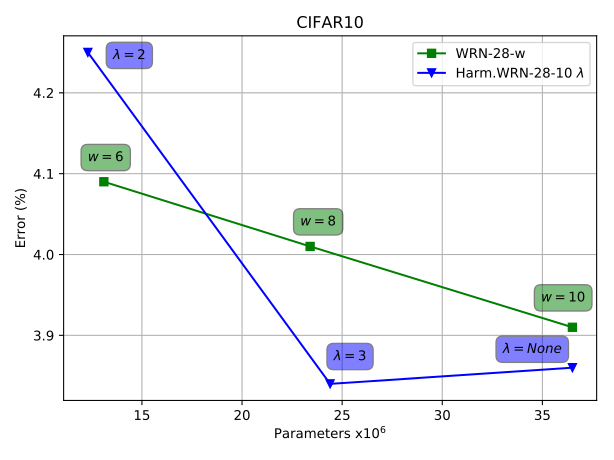}
   \includegraphics[width=.49\linewidth]{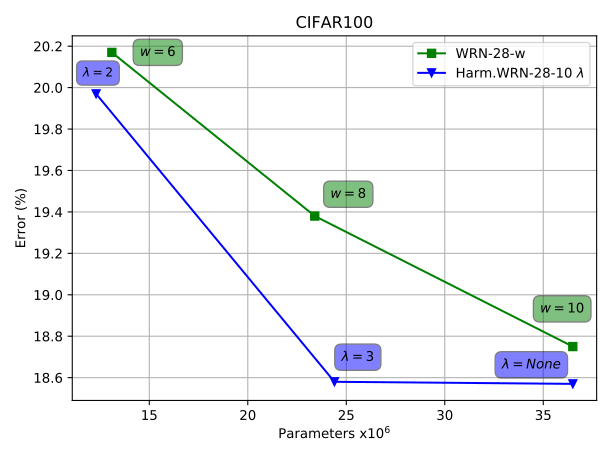}
\end{center}
\vspace{-1.25\baselineskip}
\caption{Decrease of classification error as a function of model size on CIFAR-10 (left) and CIFAR-100 (right). Parameters of harmonic networks are controlled by the compression parameter $\lambda$, the WRN baselines by the width multiplier {\it w}.}
\label{fig:compression_graph}
\end{figure*}

Analysis of fully harmonic WRN weights learned with 3x3 spectrum revealed that the deeper layers tend to favour low-frequency information over high frequencies when learning representations. Relative importance of weights corresponding to different frequencies shown in Fig.~\ref{fig:freq_heatmap} motivates truncation of high-frequency coefficients for compression purposes. While preserving the input image spectrum intact, we train the harmonic networks on limited spectrum of hidden features for $\lambda$=2 and $\lambda$=3 using 3 and 6 DCT bases respectively. 
To assess the loss of accuracy associated with parameter reduction we train baselines with reduced widths having comparable numbers of parameters: WRN-28-8 and WRN-28-6, see Fig.~\ref{fig:compression_graph}. Fully harmonic WRN-28-10 with $\lambda$=3 has comparable error to the network using the full spectrum and outperforms the larger baseline WRN-28-10, showing almost no loss in discriminatory information. On the other hand Harm-WRN-28-10 with $\lambda$=2 is better on CIFAR-100 and slightly worse on CIFAR-10 compared to the similarly sized WRN-28-6. The performance degradation indicates that some of the truncated coefficients carry important discriminatory information.

We further compare performance with the Gabor CNN 3-28~\cite{Luan18} that relies on modulating learned filters with Gabor orientation filters. To operate on a similar model we remove dropouts and reduce complexity by applying progressive $\lambda$: no compression for filters on 32x32 features, $\lambda$=3 for 16x16, and $\lambda$=2 for the rest. With a smaller number of parameters the Harm-WRN-28-10 performs similarly on CIFAR-10 and outperforms Gabor CNN on CIFAR-100.

\vskip .1cm
\noindent{\bf Harmonic block implementations.}
Here we compare the standard harmonic block implementation with its memory efficient version introduced in Algorithm~\ref{alg:mem_eff_harm_block}, see Table~\ref{tab:alg_req}. The comparison on CIFAR-10 dataset demonstrates that Algorithm~\ref{alg:mem_eff_harm_block} provides similar overall performance but reduces both the runtime and memory requirements nearly three times. We will therefore use solely this implementation of the harmonic block except for the root (first) layer due to the use of BN on that first layer.

\begin{table}[b]
\caption{Modifications of the WRN-16-4 baseline on CIFAR-100: mean classification errors and standard deviations from 5 runs when replacing particular layers by harmonic blocks.} \label{tab:ablation}
\vspace{0.3\baselineskip}
\centering
\begin{tabular}{ c c c c }
 \hline
 \textbf{Root block} & \textbf{Harmonic root BN} & \textbf{Residual blocks} & \textbf{Error \%} $\downarrow$ \\
 \hline
 & & & 24.07 $\pm$ 0.24 \\
 \checkmark & & & 23.79 $\pm$ 0.24 \\
 \checkmark & \checkmark & & 23.67 $\pm$ 0.12 \\
 & & \checkmark & 23.22 $\pm$ 0.28 \\
 \checkmark & & \checkmark & 23.25 $\pm$ 0.25 \\
 \checkmark & \checkmark & \checkmark & 23.21 $\pm$ 0.11 \\
 \hline
\end{tabular}
\end{table}

\vskip .1cm
\noindent{\bf Ablation study.}
The effect of filter parametrisation by DCT bases is investigated by replacing particular layers of WRN-16-4 (w/o dropout) with harmonic blocks, see Table~\ref{tab:ablation}. 
We consider replacing the root convolution layer (with or without BN), or layers in residual blocks. Replacing each layer has provided the greatest improvement, while BN in the first block decreases the variance by half.
These observations correspond to the results obtained on NORB dataset. We will always be employing BN as part of the root harmonic block.

\vskip .1cm
\noindent{\bf Compressing existing models.}
Section~\ref{sec:method.subsample} described how convolutional filters in certain layer can be approximated with fewer parameters. So far we have only considered uniform coefficient truncation by truncating the same frequencies in all the layers, or a simple progressive compression. In this experiment we assess the effectiveness of our more elaborate coefficient selection schemes on already trained model. We start with the WRN-28-10 baseline trained without dropout, which has been converted to harmonic WRN-28-10 net (omitting BN in the first harmonic block) by re-expressing each 3$\times$3 filter as a combination of DCT basis functions. The first harmonic block is kept intact (no compression in DCT representation), while all other blocks are compressed. We compare three different coefficient selection strategies: uniform, advanced progressive and adaptive selection  (cf. Sec \ref{sec:method.subsample}).

\begin{figure}
\begin{center}
  \includegraphics[width=.6\linewidth]{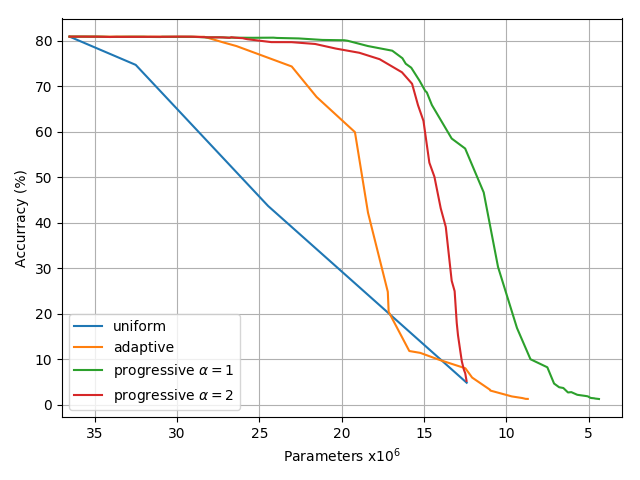}
\end{center}
\vspace{-.5\baselineskip}
\caption{Accuracy of compressed WRN-28-10 on CIFAR-100 dataset using different coefficient truncation strategies.}
\label{fig:compression_types}
\end{figure}

The results reported in Fig.~\ref{fig:compression_types} confirm the behavior observed in Fig.~\ref{fig:freq_heatmap}, i.e. the high frequencies appear to be more relevant in the early layers of the network compared to deeper layers. The uniform compression discards the same amount of information in all the layers, and is surpassed by the other compression strategies. By using progressive or adaptive coefficient selection a model can be compressed by over 20\% without a loss in accuracy. The best progressive method loses less than 1\% of accuracy when compressed by 45\% without a need for finetuning.

\subsection{ImageNet dataset} 
\label{sec:experiments.imagenet}

We present here results obtained on ImageNet-1K classification task.
ResNet~\cite{He16} with 50 layers is adopted as the baseline. 
To reduce memory consumption maxpooling is not used, instead the first convolution layer employs stride 4 to produce equally-sized features; we refer to this modification as ResNet-50 (no maxpool). The following harmonic modifications refer to this baseline without maxpooling after the first layer. We investigate the performance of three harmonic modifications of the baseline: (i) replacing solely the initial 7x7 convolution layer with harmonic block (with BN) with 7x7 DCT filters, (ii) replacing all convolution layers with receptive field larger than 1x1 with equally-sized harmonic blocks, (iii) compressed version of the fully-harmonic network.
The models are trained as described in~\cite{Ulicny19b}, and here we report accuracy after 100 epochs.

\begin{table}[h]
\caption{Classification errors on ImageNet validation set using central crops.} 
\label{tab:imagenet_spec}
\vspace{0.3\baselineskip}
\centering
\begin{tabular}{ l c c c }
\hline
 \textbf{Model} & \textbf{Parameters} $\downarrow$ & \textbf{Top-1 \%} $\downarrow$ & \textbf{Top-5 \%} $\downarrow$ \\
\hline
 VGG16-BN & 138.4M & 25.86 & 8.05 \\
 Harm-VGG16-BN & 138.4M & 25.55 & 8.01 \\
 ResNet-50 (no maxpool) & 25.6M & 23.83 & 7.01 \\
 Harm1-ResNet-50 & 25.6M & 23.01 & \textbf{6.47} \\
 Harm-ResNet-50 & 25.6M & \textbf{22.98} & 6.64 \\
 Harm-ResNet-50, progr. $\lambda$ & 19.7M & 23.21 & 6.67 \\
 Harm-ResNet-101 & 44.5M & 21.45 & 5.78 \\

\hline
{\bf Benchmarks}\\
 ResNet-50 (maxpool)* & 25.6M & 23.87 & 7.14 \\
 ScatResNet-50~\cite{Oyallon18b} & 27.8M & 25.5 & 8.0 \\
 JPEG-ResNet-50~\cite{Gueguen18} & 28.4M & 23.94 & 6.98 \\
 ResNet-101 (maxpool)* & 44.5M & 22.63 & 6.45 \\
\hline
\end{tabular}\\[.05cm]
*torchvision models.
\end{table}

\begin{table}[h]
\caption{Performance of the converted harmonic networks (error on ImageNet).} 
\label{tab:imagenet_spec2}
\vspace{0.3\baselineskip}
\small
\centering
\tabcolsep = 1.5mm
\begin{tabular}{ lclcc }
\hline
\textbf{Training} & \textbf{Epochs} & \textbf{Model} & \textbf{Top-1\%}$\downarrow$ & \textbf{Top-5\%}$\downarrow$ \\
\hline
full & 90 & ResNet-50 (no maxpool) & 24.36 & 7.33 \\
finetuned & 90+5 & ResNet-50 (no maxpool) & 24.34 & 7.30 \\
\hline

finetuned & 90+5 & ResNet-50\,$\Rightarrow$\,Harm-ResNet-50 & 24.06 & 7.12 \\
finetuned & 90+5 & ResNet-50\,$\Rightarrow$\,Harm-ResNet-50, progr. $\lambda$ & 24.62 & 7.44 \\
\hline
\end{tabular}
\end{table}

\begin{figure}[h]
\begin{center}
   \includegraphics[width=0.48\linewidth]{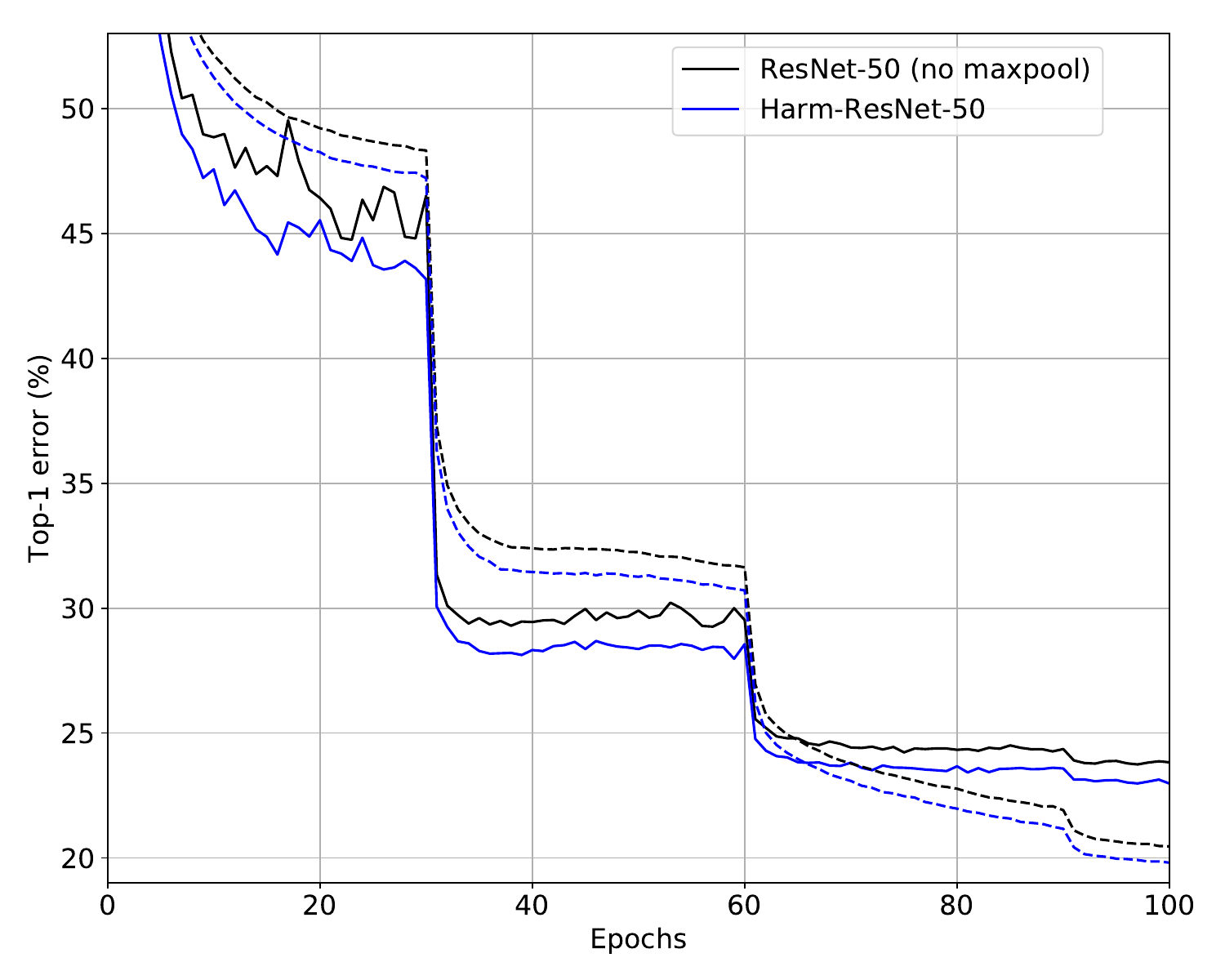}\;\;
   \includegraphics[width=0.48\linewidth]{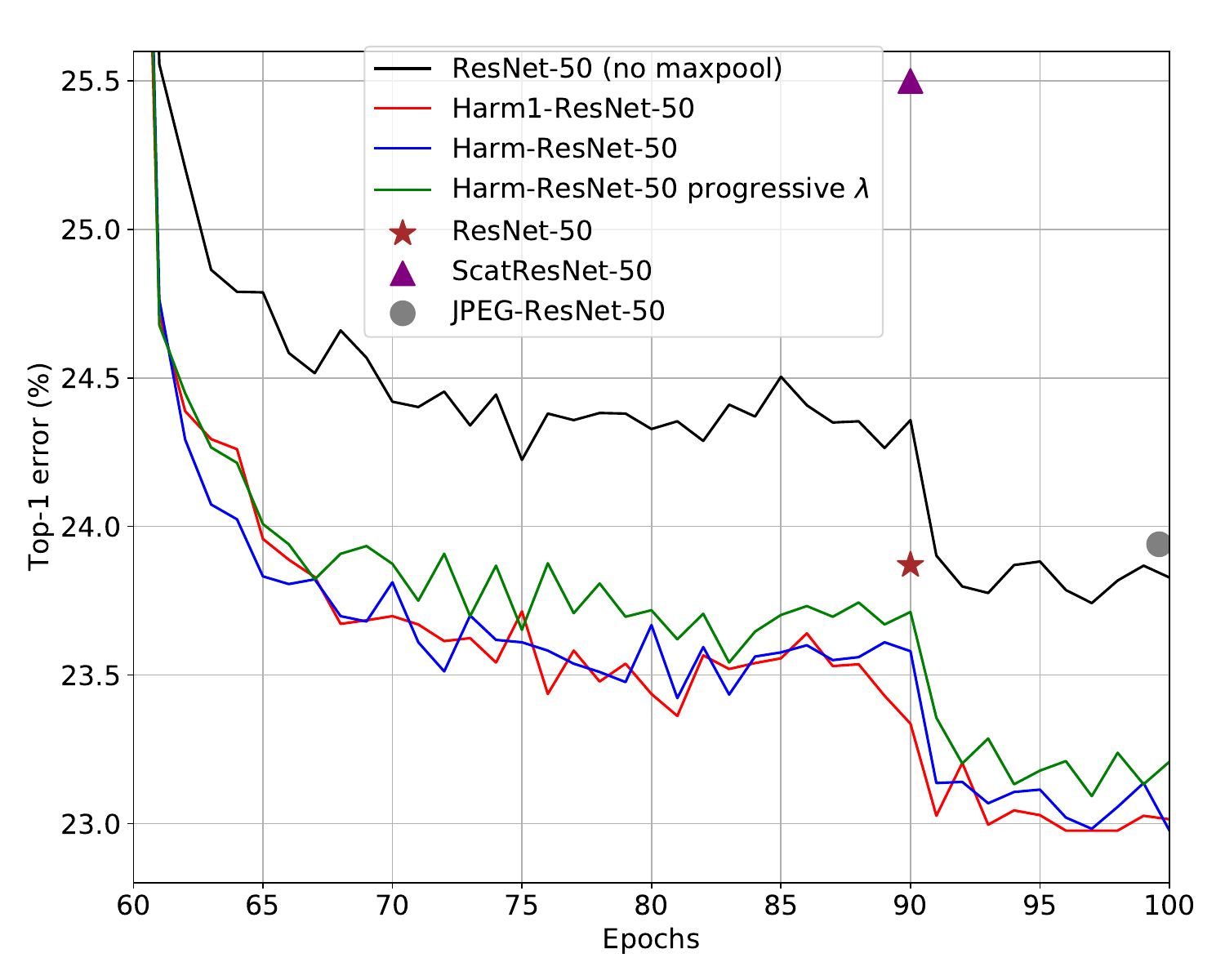}
\end{center}
\vspace{-.5\baselineskip}
   \caption{Training of harmonic networks on ImageNet classification task. Left: comparison with the baseline showing validation error (solid line) and training error (dashed). Right: last 40 epochs of training for all the ResNet-50 based models including scores reported for the benchmark models.
   }
\label{fig:imagenet_curve}
\end{figure}

Table~\ref{tab:imagenet_spec} reports error rates on ImageNet validation set using central 224$\times$224 crops from images resized such that the shorter side is 256.
All three harmonic networks have similar performance and improve over the baseline by $0.6-1\%$ in top1 and $0.4-0.6\%$ in top5 accuracy. We observe similar progress of the three modifications during training, see Fig.~\ref{fig:imagenet_curve}. ResNet-50 architecture has 17 layers with spatial filters which correspond to 11M parameters. We reduce this number by using progressive $\lambda$ compression: $\lambda$=3 on 14x14 features and $\lambda$=2 on the smallest feature maps. This reduces the number of weights roughly by half, in total by about 23\% of the network size. The compressed network loses almost no accuracy and still clearly outperforms the baseline.
Even with compression the proposed Harm-ResNet-50 confidently outperforms the standard ResNet-50 (maxpool), as well as the ScatResNet-50~\cite{Oyallon18b} and JPEG-ResNet-50~\cite{Gueguen18}.
Furthermore, we also observe a substantial improvement of 1.15 in top-1 error \% associated with the introduction of harmonic blocks into a deeper ResNet-101. 

\begin{table}[!t]
\centering
\caption{GPU training memory requirements and speed of harmonic block implementations on CIFAR-10 and ImageNet. All ImageNet models use harmonic blocks based on Alg.~\ref{alg:mem_eff_harm_block}. Values are measured on Nvidia RTX6000 using batch size 128.} \label{tab:alg_req}
\vspace{0.3\baselineskip}
 \begin{tabular}{lcccc}
  \hline
  \multirow{2}{*}{\textbf{Model}} & {\bf GPU}  & \multicolumn{2}{c}{\bf Images/s $\uparrow$}  & \multirow{2}{*}{{\bf Error \%}$\downarrow$}  \\
 &  {\bf memory$\downarrow$} & {\bf train.} & {\bf infer.} & \\
  \hline
  {\bf CIFAR-10} \\
  WRN-28-10~\cite{Zagoruyko16} & 4.6GB & 606.4 & 1876.9 & 3.89 \\
  Harm-WRN-28-10 (non-optimized) & 14.1GB & 211.0 & 600.4 & 3.71 \\
  Harm-WRN-28-10 (Alg.~\ref{alg:mem_eff_harm_block}) & 4.8GB & 573.3 & 1736.5 & 3.78 \\
  \hline
  {\bf ImageNet} \\
  ResNet-50 (no maxpool) & 11.2GB & 306.2 & 820.5  & 23.83 \\
  ResNet-50 (maxpool) & 12.1GB & 292.9 & 790.1 & 23.87 \\
  Harm-ResNet-50 & 11.4GB & 296.3 & 766.5 & 22.98 \\
  ResNet-101 (maxpool) & 17.4GB & 174.1 & 526.7 & 22.63 \\
  Harm-ResNet-101 & 16.9GB & 174.4 & 507.9 & 21.45 \\
  \hline
 \end{tabular}
\end{table}

We validate the use of harmonic blocks on an architecture without residual connections as well, specifically the VGG16~\cite{Simonyan14} architecture with BN layers. Harm-VGG16-BN, obtained by replacing all convolutional layers by harmonic blocks yields an improvement of $\sim$0.8\% in top-1 classification error. This demonstrates that the improvement is not limited to residual-based architectures. 

Finally, we evaluate conversion of weights of a pretrained non-harmonic network to those of its harmonic version. Each learned filter in the pretrained baseline (ResNet-50 without maxpooling after 90 epochs of training) is transformed into DCT domain, skiping BN inside the first harmonic block. The direct conversion resulted in the exact same numerical performance due to the basis properties of DCT.
We then finetune the converted model for another 5 epochs with the learning rate of 0.001, which results in the top1 (top5) performance improvement of 0.21\% (0.19\%) over the pretrained baseline, see Table~\ref{tab:imagenet_spec2}. We also investigate the conversion to a harmonic network with progressive $\lambda$ compression. After casting the pretrained filters into the available number of DCT filters (from full basis at the early layers to 3 out of 9 filters at the latest layers), the top1 performance degrades by 6.3\% due to loss of information. However, if we allow finetuning for as few as 5 epochs the top1 (top5) accuracy falls 0.24\% (0.09\%) short of the baseline, while reducing the number of parameters by 23\%. This analysis shows how the harmonic networks can be used to improve the accuracy and/or compress the existing pretrained CNN models.

\begin{table}[h]
\caption{SE-ResNeXt networks: harmonic vs. baseline errors and comparison with the state of the art on ImageNet.} \label{tab:resnext}
\vspace{0.3\baselineskip}
\centering
\tabcolsep = .4mm
\begin{tabular}{lccccccc}
 \hline
 \multirow{2}{*}{\textbf{Model}} & \multirow{2}{*}{\textbf{Param}} & \multicolumn{3}{c}{224$\times$224} & \multicolumn{3}{c}{320$\times$320 / 331$\times$331} \\
& & \textbf{FLOPS} & \textbf{Top-1} & \textbf{Top-5} & \textbf{FLOPS} & \textbf{Top-1} & \textbf{Top-5} \\
 \hline
 {\bf ResNeXt-101 (RNX):} \\
 RNX (64x4d)~\cite{Xie17} & 83.6M & 15.5B & 20.4 & 5.3 & 31.5B & 19.1 & 4.4 \\
 SE-RNX(32x4d) & 49.0M & 8.0B & 19.73 & 4.90 & 16.3B & 18.49 & 4.05 \\
 Harm-SE-RNX(32x4d) & 49.0M & 8.1B & 19.39 & 4.73 & 16.5B & 18.48 & 4.06 \\
 Harm-SE-RNX(64x4d) & 88.2M & 15.4B & \textbf{18.37} & \textbf{4.34} & 31.4B & \textbf{17.15} & \textbf{3.56} \\
 \hline
 {\bf Benchmarks}\\
 PolyNet~\cite{Hu19} & 92M & - & - & - & 34.7B & 18.71 & 4.25 \\
 DualPathNet-131~\cite{Hu19} & 79.5M & 16.0B & 19.93 & 5.12 & 32.0B & 18.55 & 4.16 \\
 SENet-154~\cite{Hu19} & 145.8M & 20.7B & 18.68 & 4.47 & 42.3B & 17.28 & 3.79 \\
 NASNet-A~\cite{Real19} & 88.9M & - & - & - & 23.8B & 17.3 & 3.8 \\
 AmoebaNet-A~\cite{Real19} & 86.7M & - & - & - & 23.1B & 17.2 & 3.9 \\
 PNASNet-5~\cite{Real19} & 86.1M & - & - & - & 25.0B & 17.1 & 3.8  \\
 EfficientNet-B7*~\cite{Tan19} & 66M & - & - & - & 37B & 15.6 & \textbf{2.9} \\
 Swin-B*~\cite{Liu21} & 88M & 15.4B & \textbf{16.5} & \textbf{3.5} & 47.0B & \textbf{15.5} & 3.0 \\
 \hline
\end{tabular}\\[.05cm]
*models trained on 600$\times$600 (EfficientNet) and 384$\times$384 (Swin) crops.
\end{table}

\vskip .1cm
\noindent{\bf Comparison with the cutting-edge techniques.}
Here we verify the use of DCT-based harmonic blocks in the more elaborate state-of-the-art models. To this end we modify ResNeXt architecture~\cite{Xie17}, which is similar to ResNets and uses wider bottleneck and grouped convolution to decrease the amount of FLOPS and the number of parameters. The model is further boosted using several state-of-the-art adjustments: {\it (i)} identity mappings that downsample features are extended by average pooling to prevent information loss; {\it (ii)} squeeze and excitation blocks~(SE)~\cite{Hu19} are used after every residual connection. The network is further regularized by stochastic depth and dropout on the last layer. Training is performed via stochastic gradient decent with learning rate 0.1 and batch size 256, with the former decayed according to one cosine annealing cycle. In addition to mirroring and random crops of size 224, images are augmented with rotations and random erasing. 

Our ResNeXt modification with 101 layers and 32 groups per 4 convolutional filters in a residual block is trained for 120 epochs. The use of DCT bases provides a subtle improvement over the standard bases. Furthermore, we upscale the network to use 64 groups of filters, replace max-pooling in the first layer by increased stride and train this network for 170 epochs. From Table~\ref{tab:resnext} we conclude that our model outperforms all other ``handcrafted'' architectures that do not use extra training images and performs comparably to the networks of similar complexity found via neural architecture search. Note that {these models were trained on larger image crops compared to our harmonic network, which typically leads to higher accuracies.}

Vision transformers~\cite{Liu21} process a sequence of patch embeddings and consider global relations within an image, which is distinct to the CNN approach. At the same time transformers lack some inductive biases such as translation equivariance, locality and have issues related to the features' scale. Recently proposed Swin transformers~\cite{Liu21} address some of these issues by hierarchical representation and strong augmentation to perform very well even without pre-training on larger datasets (cf. Tab. \ref{tab:resnext}). This excellent performance, surpassing that achieved by our model, is obtained by using more sophisticated augmentation techniques and more training epochs.


\section{Object Detection and Segmentation}
\label{sec:experiments.detection}

Representations learned from features expressed via harmonic basis are versatile and can serve well for transfer learning.  We demonstrate here that popular vision architectures relying on harmonic backbones provide a notable improvement in accuracy compared to the use of standard convolution-based backbone models. To this end we assess the performance of harmonic networks in object detection, instance and semantic segmentation tasks. For object detection, the popular single stage RetinaNet~\cite{Lin17b} and multistage Faster~\cite{Ren15} and Mask R-CNN~\cite{He17} frameworks are built upon our harmonic ResNet backbones. The semantic segmentation pipeline extends these backbones to DeepLabV3~\cite{Chen17rethinking} models. A set of experiments is conducted on the datasets Pascal VOC~\cite{pascal} (Sec. \ref{sec:segmentation:Pascal}) and MS COCO~\cite{mscoco} (Sec. \ref{sec:segmentation:coco}).

\begin{table}[b]
\caption{Mean average precision of Faster R-CNN models after 5 runs on Pascal VOC07 test set. ResNet-101-based models are trained once.} \label{tab:VOC}
\vspace{0.3\baselineskip}
\centering
\begin{tabular}{ l c c }
 \hline
 \textbf{Backbone} & $\uparrow$ {\textbf{Box AP VOC07}} & $\uparrow$ {\textbf{Box AP VOC07+12}} \\
\hline
ResNet-50 & 73.8 $\pm$ 0.3  & 79.7 $\pm$ 0.3 \\
Harm-ResNet-50 & \textbf{75.0 $\pm$ 0.4} & \textbf{80.7 $\pm$ 0.2} \\
\hline
ResNet-101 & 76.1 & 82.1 \\
Harm-ResNet-101 & \textbf{77.4} & \textbf{82.9} \\
 \hline
\end{tabular}
\end{table}

\subsection{Object Detection on Pascal VOC}\label{sec:segmentation:Pascal}

We extend PyTorch implementation provided by Chen et al.~\cite{mmdet} and train Faster R-CNN model based on our harmonic ResNets with 50 and 101 layers. 
Region proposal network (RPN) is applied on the feature pyramid~\cite{Lin17} constructed from the network layers. RPN layers as well as regression and classification heads are randomly initialized and use standard (non-harmonic) convolution/fully connected layers. Images are resized to set their shortest sides at 600 pixels. The Faster R-CNN is trained with the learning rate $lr=0.01\times(bs/16)$ dependant on a particular batch size $bs$. Models are trained on the union of VOC 2007 training and validation sets with about 5000 images for 17 epochs, decreasing the learning rate by a multiplicative factor of 0.1 after epoch 15. We train the networks with original and harmonic backbones using the same setting. Additionally, these models are also trained on the combination of training sets of VOC 2007 and VOC 2012, consisting of about 16 500 images, for 12 epochs with learning rate dropped at epoch 9. All models are tested on VOC 2007 test set and the official evaluation metric, the mean average precision (AP), is averaged over 5 runs. Final results are reported in Table~\ref{tab:VOC}  for different depths of ResNet backbones and configurations of the dataset.

From Table~\ref{tab:VOC} we conclude that the models built on our harmonic backbones surpass their conventional convolution-based counterparts in all configurations as well as on both training sets. 
We observe a consistent improvement due to the Harmonic architecture: by 1\% AP for ResNet-50 and 0.8\% AP in case of ResNet-101 using the Faster R-CNN architecture.

\subsection{Object Detection on MS COCO}\label{sec:segmentation:coco}
Common Objects in COntext (COCO) dataset poses a greater challenge due to a higher variety of target classes and generally smaller object sizes.
The networks are trained following the standard procedure, images resized so that their shortest side is 800 pixels. The learning rate is initialized by linear scaling method $lr=0.02\times(bs/16)$ using default hyperparameters set up by Chen et al.~\cite{mmdet}. 
All models are trained with standard 12 (24) epochs schedules with learning rate decreased by 10 after epochs 8 (16) and 11 (22). Table~\ref{tab:COCO} shows that the use of our harmonic backbones consistently improves both single-stage RetinaNet and multi-stage Faster and Mask R-CNN detectors by 0.7-1.3 AP with identical training procedures employed.

\begin{table}[t]
\caption{Mean average precision for different backbones and detector types on MS COCO 2017 validation set. All backbones are transformed to FPNs.} \label{tab:COCO}
\vspace{0.3\baselineskip}
\centering
\tabcolsep = .55mm
\begin{tabular}{ lccccc }
 \hline
 \multirow{2}{*}{\textbf{Backbone}} & \multirow{2}{*}{\textbf{Type}} & \multicolumn{2}{c}{\textbf{Box AP} $\uparrow$} & \multicolumn{2}{c}{\textbf{Mask AP} $\uparrow$} \\
 & & \textbf{12 epochs} & \textbf{24 epochs} & \textbf{12 epochs} & \textbf{24 epochs} \\
 \hline
 ResNet-50 & Faster & 36.4 & \hspace{.4ex} 37.7* & - & - \\
 Harm-ResNet-50 & Faster & \textbf{37.2} & \textbf{38.4} & - & - \\
 ResNet-50 & Retina & \hspace{.4ex} 35.6* & \hspace{.4ex} 36.4* & - & - \\
 Harm-ResNet-50 & Retina & 36.3 & 36.8 & - & - \\
 \hline
 ResNet-101 & Faster & 38.5 & 39.3 & - & - \\
 Harm-ResNet-101 & Faster & \textbf{39.7} & \textbf{40.3} & - & - \\
 ResNet-101 & Retina & \hspace{.4ex} 37.7* & \hspace{.4ex} 38.1* & - & - \\
 Harm-ResNet-101 & Retina & 39.0 & 39.2 & - & -\\
 \hline
 ResNet-50 & Mask & \hspace{.4ex} 37.3* & \hspace{.4ex} 38.5* & \hspace{.4ex} 34.2* & \hspace{.4ex} 35.1* \\
 Harm-ResNet-50 & Mask & \textbf{38.1} & \textbf{38.9} & \textbf{34.7} & \textbf{35.5} \\
 \hline
 ResNet-101 & Mask & \hspace{.4ex} 39.4* & \hspace{.4ex} 40.3* & \hspace{.4ex} 35.9* & \hspace{.4ex} 36.5* \\
 Harm-ResNet-101 & Mask & \textbf{40.7} & \textbf{41.5} & \textbf{36.8} & \textbf{37.3} \\
 \hline
\end{tabular}\\[.05cm]
*scores reported by~\cite{mmdet}.
\end{table}

The state-of-the-art detectors rely on a cascade of detection heads with progressively increasing IoU thresholds, which refines the bounding boxes and thus improves localization accuracy~\cite{Cai18}. In Table~\ref{tab:cascade},  we  report comparisons achieved with the Cascade R-CNN architecture, trained using the 20-epoch schedule suggested in~\cite{Cai18}. The use of our harmonic ResNet-101 provides a  1.0 AP improvement for object detection similar to Faster \& Mask R-CNNs, and it also improves instance segmentation AP by 0.7 (see Table~\ref{tab:cascade}). Moreover, a similar improvement of 1.1 AP is observed for   hybrid task cascade R-CNN~\cite{Chen19} that alters the mask refinement procedure and exploits semantic segmentation information to incorporate additional contextual information.

\begin{table}[t]
\caption{Mean average precision on Cascade R-CNN architecture on MS COCO 2017 validation set. All backbones are transformed to FPNs.} \label{tab:cascade}
\vspace{0.3\baselineskip}
\centering
\begin{tabular}{ lccc }
 \hline
 \textbf{Cascade R-CNN Backbone} & \textbf{Type} & \textbf{Box AP} $\uparrow$ & \textbf{Mask AP} $\uparrow$ \\
 \hline
 ResNet-101 & Faster & \hspace{.4ex} 42.5* & - \\
 Harm-ResNet-101 & Faster & 43.5 &  - \\
 \hline
 ResNet-101 & Mask & \hspace{.4ex} 43.3* & \hspace{.4ex} 37.6* \\
 Harm-ResNet-101 & Mask & 44.3 & 38.3 \\
 \hline
 ResNet-101 & Hybrid & \hspace{.4ex} 44.9* & \hspace{.4ex} 39.4* \\
 Harm-ResNet-101 & Hybrid & {\bf 46.0} & {\bf 40.2}  \\
 \hline
\end{tabular}\\[.05cm]
*scores reported by~\cite{mmdet}.
\end{table}

These experiments on object detection and localization demonstrate that the harmonic versions of the backbones provide a meaningful improvement of about 1.0 AP in terms of both bounding boxes and masks to the state-of-the-art detection architectures. Our harmonic networks retain this improvement from the purely classification task through the transformation to the Feature Pyramid Networks (FPNs).

\subsection{Semantic Segmentation on Pascal VOC} \label{sec:experiments.semantic}

\begin{figure}[!t]
\centering
   \includegraphics[width=0.24\linewidth]{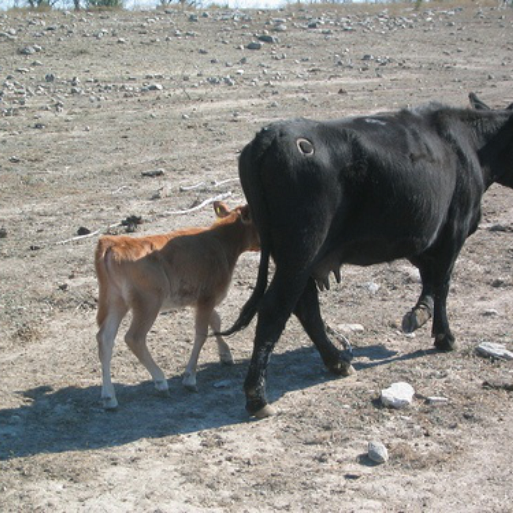}
   \includegraphics[width=0.24\linewidth]{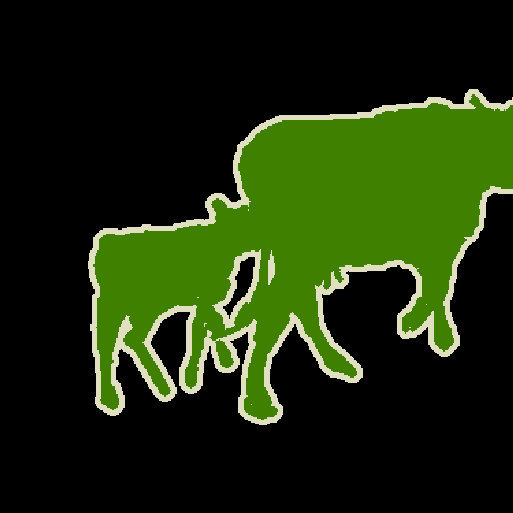}
   \includegraphics[width=0.24\linewidth]{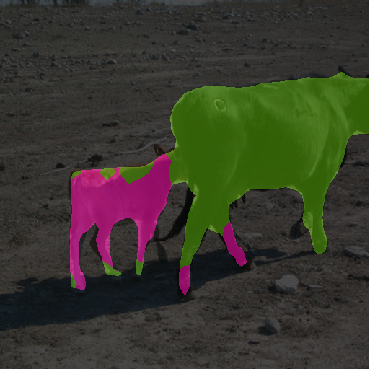}
   \includegraphics[width=0.24\linewidth]{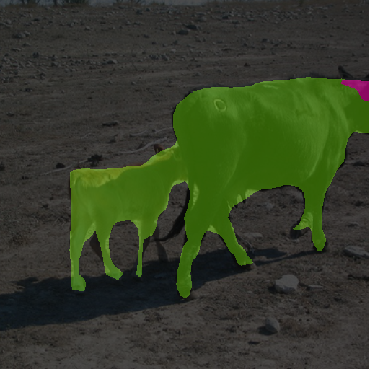}\\[.05cm]
   \includegraphics[width=0.24\linewidth]{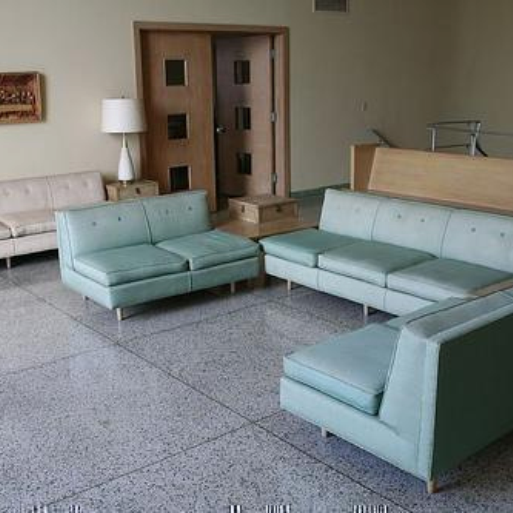}
   \includegraphics[width=0.24\linewidth]{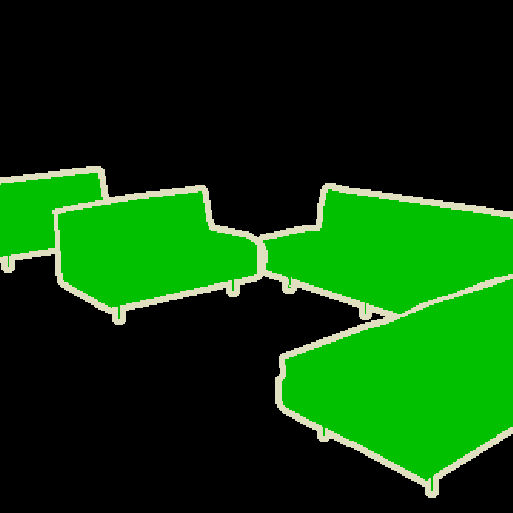}
   \includegraphics[width=0.24\linewidth]{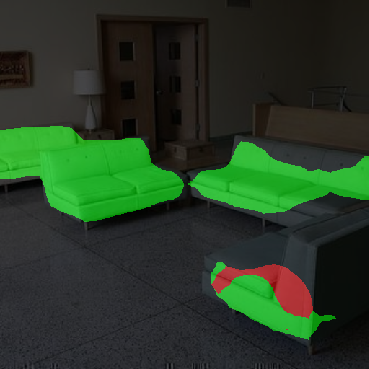}
   \includegraphics[width=0.24\linewidth]{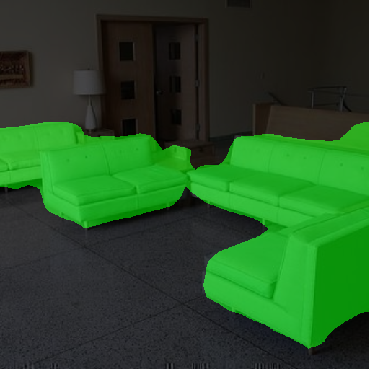}\\[.05cm]
   \includegraphics[width=0.24\linewidth]{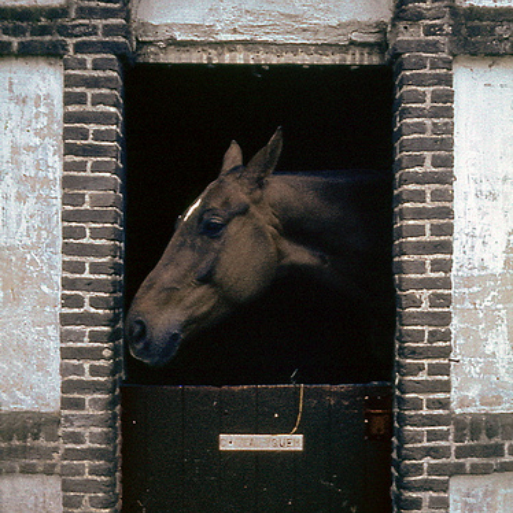}
   \includegraphics[width=0.24\linewidth]{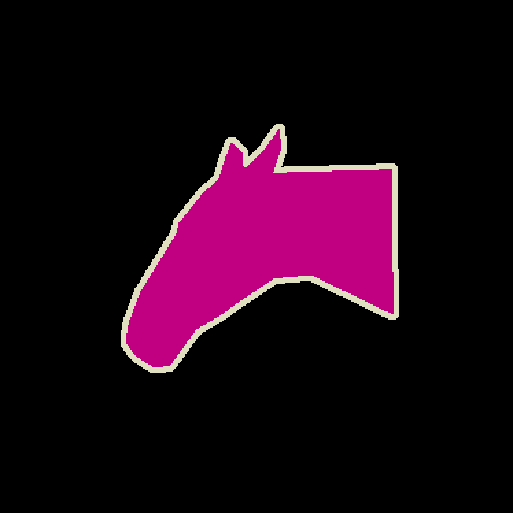}
   \includegraphics[width=0.24\linewidth]{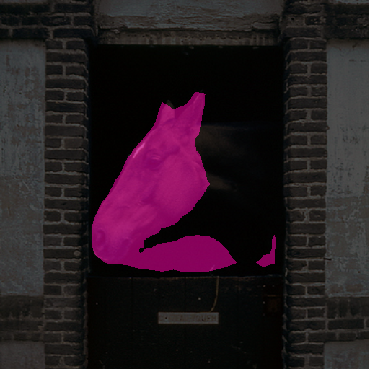}
   \includegraphics[width=0.24\linewidth]{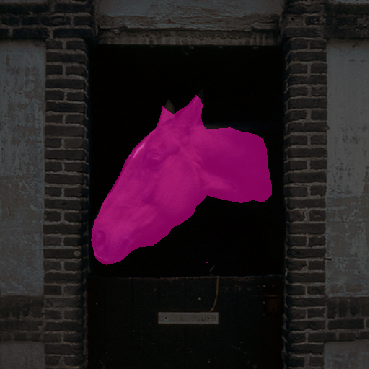}\\[.05cm]
   \includegraphics[width=0.24\linewidth]{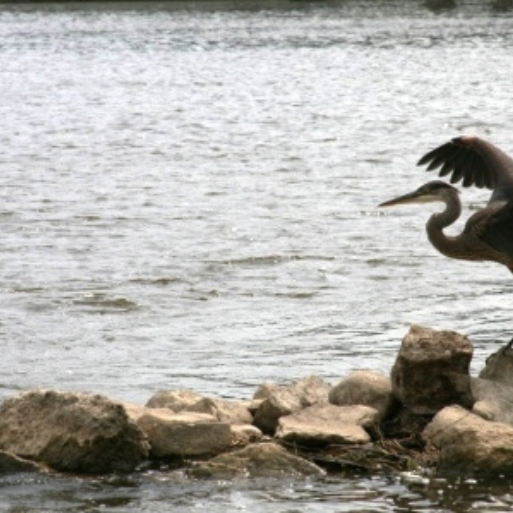}
   \includegraphics[width=0.24\linewidth]{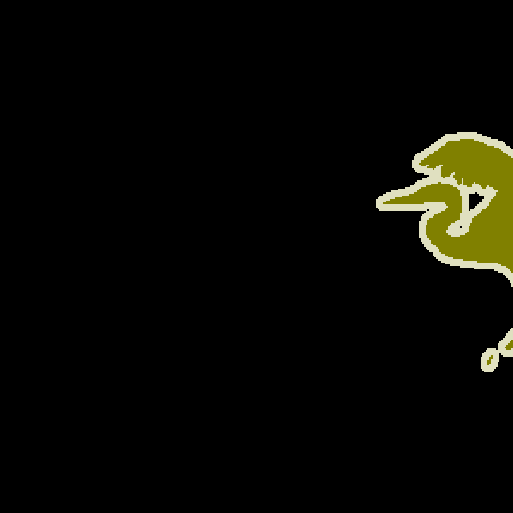}
   \includegraphics[width=0.24\linewidth]{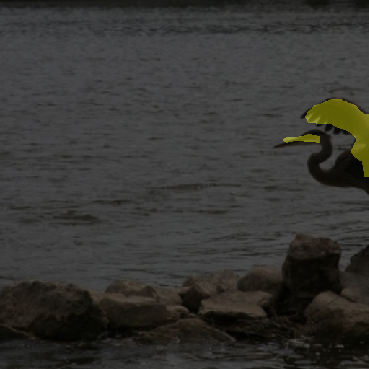}
   \includegraphics[width=0.24\linewidth]{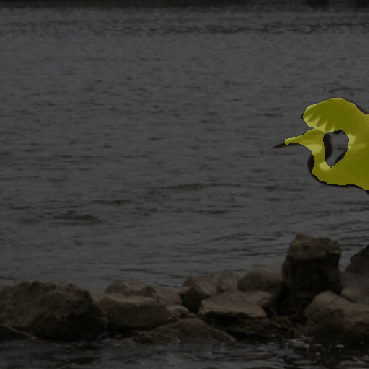}\\[.05cm]
   \includegraphics[width=0.24\linewidth]{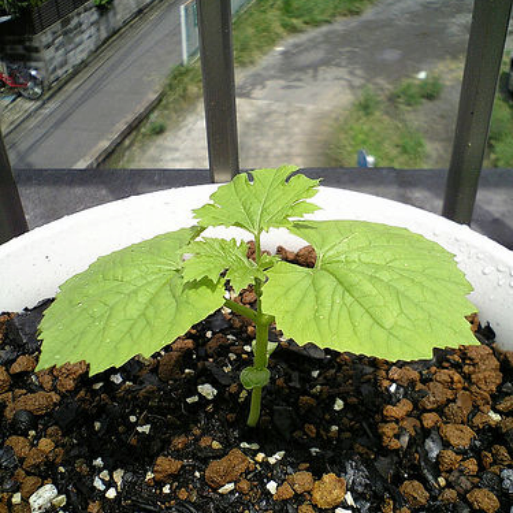}
   \includegraphics[width=0.24\linewidth]{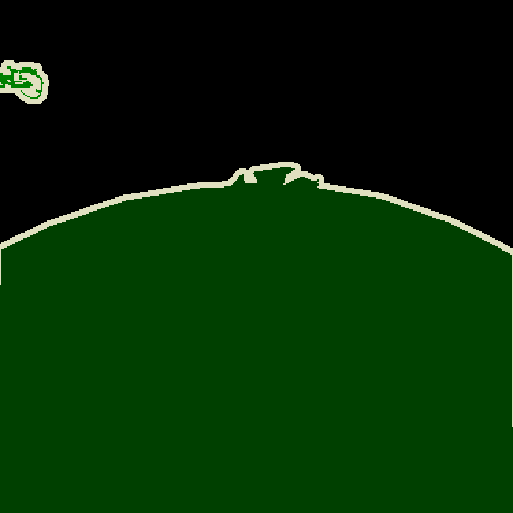}
   \includegraphics[width=0.24\linewidth]{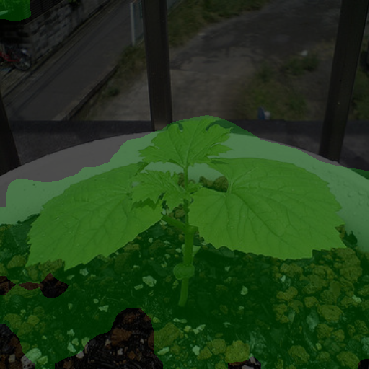}
   \includegraphics[width=0.24\linewidth]{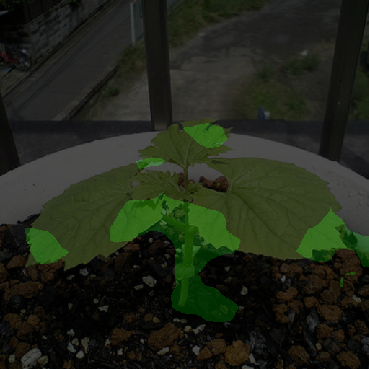}\\[-.25cm]
   \setlength{\unitlength}{1cm}
   \begin{picture}(12,0.65)
     \put(0,0){Original}
     \put(3.2,0){Annotation}
     \put(6.9,0){DeepLabV3}
     \put(10.0,0){Harm-DeepLabV3}
   \end{picture}
\vspace{-.5\baselineskip}
   \caption{\small Examples of semantic segmentation on Pascal VOC 2012 validation images. The first 4 rows show where harmonic network is more successful than the baseline, while the last row displays case where it fails. DeepLabV3 with ResNet-101 backbone is used.}
\label{fig:segmentations}
\end{figure}

We now assess our proposed harmonic networks on the task of semantic segmentation using the Pascal VOC 2012 benchmark. Training images are augmented into a set of 10,582 samples as in~\cite{Chen17rethinking}. Performance is measured in terms of intersection over union (IoU) on a large validation set consisting of 1449 images. The segmentation is performed using the DeepLabV3 architecture~\cite{Chen17rethinking}. We extend  PyTorch implementation of this model\footnote{\url{https://github.com/VainF/DeepLabV3Plus-Pytorch}} and retrain baseline models with ResNet-50 and ResNet-101 backbones for 30,000 iterations with batch size 16, learning rate 0.1 and output stride parameter equal to 16.
We replace the backbone model of the segmentation network with harmonic ResNets using two settings for the backbone: (i) converting from the original backbone models or (ii) taking a harmonic models pre-trained on ImageNet (90~epochs), see Table~\ref{tab:imagenet_spec}. The results are summarized in Table~\ref{tab:semantic}. The DeepLabV3 models with harmonic backbones pretrained on ImageNet improve IoU scores by approximately 1.1\%. This experiment also validates the application of harmonic blocks with dilated convolution, which is dissimilar to the classical dense formulation in that the spatial correlation patterns may be weaker due to dilatation. 
DeepLabV3 with harmonic backbone has the strongest average improvement on classes ``chair'' (+9.5\%), ``sheep'' (+3.9\%), ``boat'' (+3.2\%), ``cow'' (+2.9\%) and ``tvmonitor'' (+2.9\%), and performs worse on ``pottedplant'' (-3.4\%) and ``aeroplane'' (-0.8\%). 
A selection of image samples from these classes is presented in Fig.~\ref{fig:segmentations} showing how harmonic networks impact image segmentation quality. We observe some non-trivial improvements of the segmentation masks due to the use of harmonic blocks.

\begin{table}[h]
\caption{Intersection over Union (IoU) of DeepLabV3 architecture semantic segmentation on Pascal VOC 2012 validation set. IoU shown is the median of 5 trials $\pm$ empirical std. dev.} \label{tab:semantic}
\vspace{0.3\baselineskip}
\centering
\begin{tabular}{ lcccc }
 \hline
 \multirow{2}{*}{\textbf{Backbone}} & \multirow{2}{*}{\textbf{Baseline}} & \textbf{Harm} & \textbf{Harm} & \multirow{2}{*}{\textbf{Benchmark}} \\
 & & \textbf{converted} & \textbf{pre-trained} & \\
 \hline
 ResNet-50 & 76.31 $\pm$ 0.07 & 76.65 $\pm$ 0.07 & \textbf{77.40 $\pm$ 0.08} & - \\
 ResNet-101 & 78.31 $\pm$ 0.07 & 77.92 $\pm$ 0.11 & \textbf{79.49 $\pm$ 0.29} & 77.21~\cite{Chen17rethinking} \\
 \hline
\end{tabular}
\end{table}

\section{Conclusion} \label{sec:conclusion}

We have presented a novel approach to explicitly incorporate spectral information extracted via DCT into CNN models. 
We have empirically evaluated the use of our harmonic blocks with the well-established state-of-the-art CNN architectures, and shown that our approach  improves results for a range of applications including image classification (0.7-1.2\% accuracy on ImageNet), object detection (0.7-1.1 AP on Pascal VOC and MS COCO) and semantic segmentation (1.1\% IoU on Pascal VOC).
We further establish that the memory footprint of harmonic nets is similar and the computational complexity increases only slightly when compared to the standard convolutional baseline architectures.
We ascertain that harmonic networks can be efficiently set-up by converting the pretrained CNN baselines. 
The use of DCT allows one to order the harmonic block parameters by their significance from the most relevant low frequency to less important high frequencies. This enables efficient model compression by parameter truncation with only minor degradation in the model performance. Current efforts aim at investigating robustness of harmonic networks and at compressing weights according to correlations across filters in depth direction.

\section*{Acknowledgements}
This research was conducted with the financial support of Science Foundation Ireland under Grant Agreement No. 13/RC/2106 and 13/RC/2106\_P2 at the ADAPT SFI Research Centre at Trinity College Dublin, Maynooth University and Dublin City University. ADAPT, the SFI Research Centre for AI-Driven Digital Content Technology, is funded by Science Foundation Ireland through the SFI Research Centres Programme.
We gratefully acknowledge the support of NVIDIA Corporation with the donation of GPUs.

\bibliography{biblio}

\end{document}